\documentclass[]{ceurart}

\usepackage{listings}
\usepackage[T1]{fontenc}
\usepackage{graphicx}
\usepackage{booktabs}
\usepackage{amsmath,amssymb,amsfonts}
\usepackage{array}
\usepackage{url}
\usepackage{subcaption}
\usepackage{tikz}

\usetikzlibrary{calc,arrows,positioning,shapes,decorations.pathmorphing}
\usepackage{pgfplots}
\pgfplotsset{compat=1.18}

\begin{document}

\copyrightyear{2026}
\copyrightclause{Copyright for this paper by its authors. Use permitted under Creative Commons License Attribution 4.0 International (CC BY 4.0).}
\conference{Graph-enhanced LLMs for trustwOrthy Web data management (GLOW), in ISWC 2026 Workshops Joint Proceedings, October 25--26, 2026, Bari, Italy}

\title{HANIA: Planner-Guided Multimodal Graph Evidence Selection for Grounded Question Answering}

\author[1]{Zafar Ali}[email=zafar_ali@seu.edu.cn, ] 
\author[1]{Asad Khan}[email=asadkhan@seu.edu.cn, ] 
\author[2]{Nimbeshaho Thierry}[email=thierryn@pdx.edu, ] 
\author[1]{Nabila Amir}[email=nabilaameer1@gmail.com, ] 
\author[4]{Adam A. Q. Mohammed}[email=adammohammed@sdu.edu.cn, ] 
\author[3]{Pavlos Kefalas}[email=pavloskefalas@gmail.com, ] 
\cormark[1] 

\address[1]{School of Computer Science and Engineering, Southeast University, Nanjing, China} 
\address[2]{Portland Institute Nanjing, NJUPT, Nanjing, China} 
\address[3]{Dashub \& School of Informatics, Aristotle University of Thessaloniki, Greece} 
\address[4]{School of Information Science and Engineering, Shandong University, Qingdao, China} 

\cortext[1]{Corresponding author.}

\begin{abstract}
Multimodal question answering remains sensitive to noisy, incomplete, and weakly grounded evidence. Long unstructured contexts can introduce redundancy and encourage unsupported generation, while flat retrieval may overlook relations needed for multi-step reasoning. We present \textbf{HANIA}, a planner-guided multimodal graph framework for evidence-grounded question answering. HANIA processes the supplied image and text using a frozen vision--language model to extract concise question-relevant visual evidence with explicit abstention. It then constructs an input-grounded multimodal graph and applies a two-group finite-state planner to coordinate descriptive and relational evidence. Coverage-aware pruning retains a compact evidence set based on relevance, graph confidence, concept coverage, and modality diversity. The selected passages, visual statements, and graph triples are provided to a frozen instruction-tuned decoder. We evaluate HANIA on \textsc{ScienceQA} using answer accuracy, evidence-filtering quality, evidence-budget sensitivity, and efficiency. The results show that structured evidence planning and compact graph-guided retrieval can support competitive multimodal question answering without target-dataset fine-tuning or iterative retrieval. The code is available at \url{https://github.com/Zafar-southeast/HANIA}. 
\end{abstract}


\maketitle
\section{Introduction}
\label{sec:Intro}

Multimodal question answering requires grounded reasoning over heterogeneous evidence, including text, images or diagrams, and relational knowledge. Knowledge-graph question answering (KG-QA)~\cite{jiangunikgqa} and retrieval-augmented generation (RAG)~\cite{chen2022murag} offer complementary strengths: graphs expose relations useful for multi-step reasoning, while retrieval helps identify relevant evidence. However, multimodal KG-assisted QA remains challenging under limited computational budgets because visual and textual information must be grounded, connected, and organized before generation. Extracted triples may be weakly aligned with visual content because of missing or misgrounded relations~\cite{zheng2023mmkgr}, while vision--language models can still produce unsupported answers even when relevant evidence is available~\cite{Leng_2024_CVPR}.

Graph-aware RAG methods address part of this problem by organizing evidence through explicit relational structure. KG$^2$RAG performs graph-guided expansion and chunk organization~\cite{zhu2025kg2rag}, while KERAG combines subgraph retrieval with filtering and summarization~\cite{sun2025kerag}. Other approaches introduce explicit reasoning control, including PMR for progressive KGQA~\cite{ren2026pmr}, SKRAG with finite-state reasoning skeletons~\cite{xu-etal-2025-skrag}, and GLOW with GNN--LLM integration for open-world KGQA~\cite{abdallah-etal-2026-leveraging}. Nevertheless, many existing methods remain text-centric, assume a pre-existing symbolic graph, or rely on complex training and retrieval pipelines. They provide limited support for constructing compact, question-aware graph evidence directly from supplied multimodal inputs.

HANIA addresses this gap through input-grounded graph construction and planner-guided evidence control. Given a question and its supplied image and text, \textsc{Qwen3-VL-Instruct}~\cite{bai2025qwen3vl} extracts concise question-relevant visual evidence with explicit abstention. Controlled rules and \textsc{GLiREL}~\cite{boylan2025glirel} then extract question-aware relations from the visual and textual evidence. The resulting input-grounded graph preserves links between each retained statement or triple and its original source.

A two-group finite-state planner coordinates descriptive and relational evidence requirements. Coverage-aware pruning then retains a compact evidence set based on question relevance, graph confidence, concept coverage, and modality diversity. The selected passages, visual statements, and graph triples are serialized for a frozen \textsc{Qwen3-8B-Instruct} decoder~\cite{yang2025qwen3}. HANIA therefore controls the evidence supplied to the decoder without target-dataset fine-tuning, decoder modification, external knowledge retrieval, or iterative retrieval. The main contributions are:

\begin{itemize}
\item We introduce \textbf{HANIA}, a planner-guided multimodal graph framework that constructs question-aware evidence directly from supplied images and text.
\item We build an input-grounded multimodal graph with explicit provenance, preserving the connection between extracted relations and their supporting evidence.
\item We propose a \textbf{two-group finite-state planner} with coverage-aware pruning and evaluate its effectiveness on \textsc{ScienceQA} using answer accuracy, evidence-filtering quality, evidence-budget sensitivity, and efficiency.
\end{itemize}

\section{Related Work}
\label{sec:related}

Research on question answering has developed along four closely related directions: KG-based reasoning, multimodal QA, retrieval-augmented generation, and structured LLM reasoning. We briefly review these areas and position \textbf{HANIA} as a lightweight, planner-guided framework for constructing and selecting compact multimodal graph evidence.

KG-QA methods use structured triples to support interpretable reasoning~\cite{liu2023knowledge}. Earlier systems relied on embedding-based retrieval and semantic parsing but often struggled with unseen entities, missing relations, and scalability. Recent approaches increasingly combine graph retrieval or structured reasoning with LLMs, including PMR~\cite{ren2026pmr}, SKRAG~\cite{xu-etal-2025-skrag}, and GLOW~\cite{abdallah-etal-2026-leveraging}. These methods improve multi-hop reasoning through structured intermediate steps, but they are largely text-centric and commonly assume an existing symbolic graph. In contrast, \textbf{HANIA} constructs an instance-level multimodal graph directly from the supplied visual and textual evidence.

Multimodal QA requires coordinated reasoning over images, text, and relational information. Multimodal knowledge graph methods investigate how heterogeneous signals can support grounded reasoning~\cite{zheng2023mmkgr}. MR-MKG applies graph neural message passing over multimodal nodes~\cite{lee2024multimodal}, while VaLiK studies automated vision-to-MMKG construction with alignment-noise mitigation~\cite{liu2025aligning}. VisDoM highlights the value of structured retrieval for visually rich inputs such as tables and diagrams~\cite{suri2024visdom}, whereas VDocRAG preserves visual layout through image-based retrieval and modeling~\cite{tanaka2025vdocrag}. However, many systems still depend on coarse evidence aggregation or long multimodal contexts. \textbf{HANIA} instead converts the supplied context into compact question-aware graph evidence and organizes it through constrained planning. RAG improves factual consistency by grounding LLM outputs in retrieved evidence. Multimodal and graph-enhanced variants incorporate multimodal alignment~\cite{chen2025seeing,chen2022murag} and graph-based retrieval or organization, including KG$^2$RAG~\cite{zhu2025kg2rag}, KERAG~\cite{sun2025kerag}, and Pythia-RAG~\cite{ali2025pythia}. More recent multimodal graph systems emphasize selective evidence control: GraphLoom uses reliability-calibrated graph evidence routing~\cite{ali2026graphloom}, while MKG-Attention applies budget-aware multimodal graph retrieval and evidence control for VideoQA~\cite{khan2026mkgattention}. Instruction tuning and Chain-of-Thought can improve reasoning control~\cite{liu2023logicot}, while knowledge-augmented multimodal CoT provides structured intermediate reasoning~\cite{mondal2024kam}. \textbf{HANIA} differs by combining input-grounded multimodal graph construction, descriptive--relational planning, and coverage-aware pruning to produce a compact provenance-preserving evidence package for a frozen decoder, without external knowledge retrieval, decoder modification, or iterative retrieval.

\begin{figure}[t]
\centering
\includegraphics[width=\linewidth]{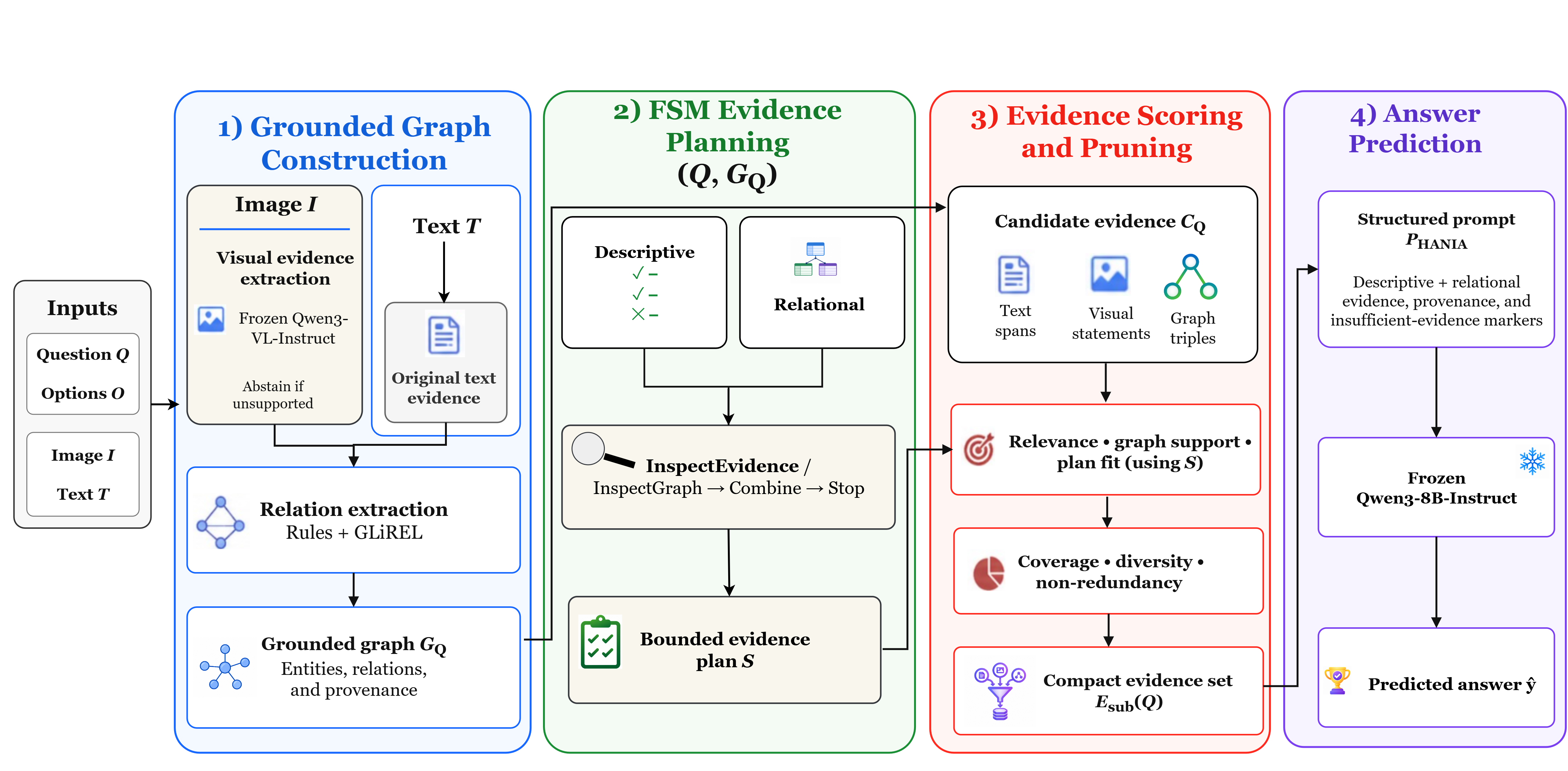}
\caption{\textbf{Architecture of HANIA.} Input-grounded multimodal evidence is organized through finite-state planning and coverage-aware pruning before prediction by a frozen decoder.}
\label{fig:proposed}
\end{figure}

\section{Methodology: HANIA}
\label{sec:Model}

Figure~\ref{fig:proposed} presents \textbf{HANIA}, a planner-guided multimodal graph framework for evidence-grounded question answering. HANIA extracts question-relevant visual and textual evidence, constructs an input-grounded graph with explicit provenance, and uses a two-group finite-state planner to coordinate descriptive and relational requirements. Coverage-aware pruning then selects a compact set of passages, visual statements, and graph triples for a frozen \textsc{Qwen3-8B-Instruct} decoder. All pretrained components are frozen, with no external knowledge retrieval.

\subsection{Input-Grounded Evidence and Graph Construction}
\label{subsec:mmkg}

Let \(Q\) denote the question, \(\mathcal{O}\) the answer options, and \(X=(I,T)\) the supplied image and textual context, where either modality may be absent. For an available image, a frozen \textsc{Qwen3-VL-Instruct} model extracts a concise, question-relevant statement:
\begin{equation}
D_Q^{I}
=
f_{\mathrm{VL}}\!\left(I,Q,\mathcal{O}\right).
\label{eq:visual_evidence}
\end{equation}
The model reports only visibly supported information and returns an abstention statement when the requested evidence cannot be determined. The original text \(T\) is retained without generative rewriting. HANIA then constructs a question-aware graph $G_Q=(V_Q,E_Q,\pi_Q),$ where \(V_Q\) contains entities, properties, and numerical values from \(D_Q^{I}\) and \(T\), \(E_Q\) contains the retained relations, and \(\pi_Q\) links each graph element to its supporting visual statement or text span.

A question-aware relation vocabulary \(\mathcal{R}_Q\) is selected based on the required evidence type, including attributes, counts, comparisons, and spatial relations. Deterministic templates convert regular visual statements into triples, while \textsc{GLiREL}~\cite{boylan2025glirel} extracts relations from textual evidence:
\begin{equation}
(\hat{r}_{jk},c_{jk})
=
f_{\mathrm{rel}}
\!\left(e_j,e_k,s;\mathcal{R}_Q\right),
\label{eq:glirel_relation}
\end{equation}
where \(s\) is the supporting evidence unit and \(c_{jk}\) is the relation confidence. A relation is retained only if its entities are valid, \(\hat{r}_{jk}\in\mathcal{R}_Q\), \(c_{jk}\geq\tau_r\), and it is relevant to \(Q\). Deterministic relations receive unit confidence. No external commonsense triples are added; unsupported requirements remain unresolved for the planning stage.

\subsection{FSM-Constrained Two-Group Planning}
\label{subsec:skeleton}

Given \(Q\) and the input-grounded graph \(G_Q\), HANIA builds a bounded evidence plan before pruning. The required evidence is organized into two groups:
\begin{equation}
\mathcal{P}
=
\{\textsc{Descriptive},\textsc{Relational}\}.
\label{eq:planner_groups}
\end{equation}
The descriptive group covers attributes, colors, materials, counts, and shapes, while the relational group covers spatial, comparative, and object-level relations.

The plan is an ordered sequence \(\mathfrak{S}=(s_1,\ldots,s_L)\), where \(s_h=(a_h,g_h,c_h)\) and \(L\leq H\). Here, \(a_h\) is the planner action, \(g_h\in\mathcal{P}\) is the active evidence group, and \(c_h\) contains relevant entity, relation, or property cues. The action space is:
\begin{equation}
\mathcal{A}_{\mathrm{plan}}
=
\{
\textsc{InspectEvidence},
\textsc{InspectGraph},
\textsc{Combine},
\textsc{Stop}
\}.
\label{eq:planner_actions}
\end{equation}

At each step, a frozen \textsc{Qwen3-8B-Instruct} planner~\cite{yang2025qwen3} proposes the next state under finite-state constraints:
\begin{equation}
s_h
=
\mathrm{Plan}_{\mathrm{FSM}}
\!\left(Q,G_Q,s_{<h}\right),
\qquad
(a_{h-1},a_h)\in\mathcal{T}_{\mathrm{FSM}},
\label{eq:fsm_planner}
\end{equation}
where \(\mathcal{T}_{\mathrm{FSM}}\) defines valid transitions. The controller rejects invalid transitions, avoids repeated inspection of the same group, and stops after at most \(H\) steps. Simple questions may use one evidence group, while more complex questions may combine both. The planner identifies evidence requirements but does not generate the final answer.

\subsection{Planner-Guided Retrieval and Coverage-Aware Pruning}
\label{subsec:retrieval}

The candidate pool \(\mathcal{C}_Q\) contains textual spans, visual statements, and graph triples, each linked to its source through \(\pi_Q\). Each candidate \(e\in\mathcal{C}_Q\) is scored as:
\begin{equation}
R(e)
=
\alpha_1 S_{\mathrm{q}}(e,Q)
+
\alpha_2 S_{\mathrm{graph}}(e)
+
\alpha_3 S_{\mathrm{plan}}(e,\mathfrak{S}),
\qquad
\sum_{i=1}^{3}\alpha_i=1,
\label{eq:evidence_score}
\end{equation}
where \(S_{\mathrm{q}}\) measures question relevance, \(S_{\mathrm{graph}}\) uses retained relation confidence or direct source support, and \(S_{\mathrm{plan}}\) rewards alignment with an active planner group. Scores are normalized to \([0,1]\), and no scorer is fine-tuned on \textsc{ScienceQA}.

A fixed top-\(K\) ranking may retain redundant evidence while missing required concepts. HANIA therefore applies greedy coverage-aware pruning. Given the selected set \(\mathcal{A}\), the marginal value of a candidate is:
\begin{equation}
\begin{aligned}
\Delta(e\mid\mathcal{A})={}
R(e)
+\beta_1\Delta_{\mathrm{cov}}(e\mid\mathcal{A})
+\beta_2\Delta_{\mathrm{div}}(e\mid\mathcal{A})
-\beta_3\mathrm{Redundancy}(e,\mathcal{A}),
\end{aligned}
\label{eq:marginal_evidence}
\end{equation}
where \(\Delta_{\mathrm{cov}}\) rewards newly covered question concepts and planner groups, while \(\Delta_{\mathrm{div}}\) rewards complementary modalities or evidence types. Starting from \(\mathcal{A}_0=\varnothing\), HANIA selects:
\begin{equation}
e_t^{*}
=
\arg\max_{e\in\mathcal{C}_Q\setminus\mathcal{A}_{t-1}}
\Delta(e\mid\mathcal{A}_{t-1}),
\qquad
\mathcal{A}_t
=
\mathcal{A}_{t-1}\cup\{e_t^{*}\}.
\label{eq:greedy_selection}
\end{equation}
Selection stops when the budget \(N_{\max}\) is reached or the best remaining score falls below \(\tau\). The final package is \(\mathcal{E}_{\mathrm{sub}}(Q)=\mathcal{A}_T\), where \(T\) is the stopping step. If a required planner group remains unsupported, HANIA inserts an insufficient-evidence marker instead of retaining weak evidence.

\subsection{Structured Evidence Prompting and Answer Generation}
\label{subsec:serialization_answer}

HANIA serializes the retained evidence \(\mathcal{E}_{\mathrm{sub}}(Q)\) while preserving the provenance links defined by \(\pi_Q\). Evidence is grouped according to the descriptive and relational requirements identified by the planner:
\begin{equation}
\mathcal{F}^{(g)}
=
\mathrm{Serialize}
\left(
\left\{
(e,\pi_Q(e))
\mid
e\in\mathcal{E}_{\mathrm{sub}}(Q),
\ \mathrm{group}(e)=g
\right\}
\right),
\quad
g\in\{\mathrm{desc},\mathrm{rel}\}.
\label{eq:group_serialization}
\end{equation}
Each item contains its original text or visual statement, any associated graph triple, and its supporting modality. If reliable support is missing, an insufficient-evidence marker is inserted. For answer options \(\mathcal{O}=\{o^{(j)}\}_{j=1}^{n}\), the structured decoder prompt is:
\begin{equation}
P_{\mathrm{HANIA}}
=
\left[
Q;
\mathcal{O};
\mathcal{F}^{(\mathrm{desc})};
\mathcal{F}^{(\mathrm{rel})}
\right].
\label{eq:hania_prompt}
\end{equation}

A frozen \textsc{Qwen3-8B-Instruct} decoder~\cite{yang2025qwen3} selects an answer using only the supplied evidence. Let \(\{y_i^{(j)}\}_{i=1}^{L_j}\) denote the tokens of option \(o^{(j)}\). Each option is scored using length-normalized log-likelihood:
\begin{equation}
S\!\left(o^{(j)}\right)
=
\frac{1}{L_j}
\sum_{i=1}^{L_j}
\log p_{\theta}
\left(
y_i^{(j)}
\mid
y_{<i}^{(j)},P_{\mathrm{HANIA}}
\right),
\qquad
\hat{y}
=
\arg\max_j S\!\left(o^{(j)}\right).
\label{eq:option_score}
\end{equation}
This constrained formulation avoids unrestricted generation and ensures that the prediction matches one of the provided options.

\subsection{Frozen-Backbone Configuration}
\label{subsec:frozen_config}

HANIA keeps the vision--language, relation-extraction, planner, and answer models frozen; none is fine-tuned on \textsc{ScienceQA}. Its contribution therefore lies in question-aware evidence construction, provenance-preserving graph representation, finite-state planning, and coverage-aware pruning. The relation-confidence threshold, planner horizon, evidence budget, prompts, and scoring weights are selected on the validation split and fixed before testing. Gold answers are used only for validation and evaluation and remain unavailable during graph construction, planning, pruning, and prediction.

\section{Experimental Evaluation}
\label{sec:Evaluation}

We evaluate \textbf{HANIA} on \textsc{ScienceQA} using answer accuracy, evidence-filtering quality, evidence-budget sensitivity, and end-to-end latency. All hyperparameters are selected on the validation split and fixed before test evaluation.

\subsection{Dataset and Evaluation Protocol}
\label{subsec:datasets_protocol}

\textsc{ScienceQA}~\cite{Lu2022ScienceQA} is a multimodal multiple-choice benchmark whose instances may contain an image, textual context, both, or neither. We use the official test split and only the supplied context, without external knowledge retrieval. Answer accuracy is the primary metric; evidence precision, recall, and F1 use candidate-level binary reference relevance labels fixed across all re-run methods and used only for evaluation. We also vary the evidence budget \(k\) and report end-to-end latency. All Table~\ref{tab:scienceqa_results} baselines are our re-run implementations on identical test instances and matched inference conditions. With deterministic decoding, each system is run once; 95\% confidence intervals use 10,000 paired bootstrap resamples of test instances. McNemar's test with Holm correction is used for the pre-specified comparisons with Qwen3-VL Direct, Flat-RAG, and SelF-Reasoner.

\subsection{Overall Results and Component Ablation}
\label{subsec:overall_results}

Table~\ref{tab:scienceqa_results} reports the controlled comparison and component-level ablations on \textsc{ScienceQA}. All methods use the same evaluation protocol, and evidence-filtering metrics are computed against the same relevance annotations.

\begin{table*}[t]
\centering
\caption{Controlled comparison and component-level ablation results on
\textsc{ScienceQA}. All values are percentages.}
\label{tab:scienceqa_results}
\resizebox{\textwidth}{!}{
\begin{tabular}{lcccc}
\hline
\rowcolor{gray!15}
\textbf{Model/Variant} &
\textbf{Accuracy} $\uparrow$ &
\textbf{Precision} $\uparrow$ &
\textbf{Recall} $\uparrow$ &
\textbf{F1} $\uparrow$ \\
\hline
\multicolumn{5}{l}{\textit{Baseline comparison}} \\
Qwen3-VL Direct~\cite{bai2025qwen3vl}
& 85.40 & -- & -- & -- \\
FOCUS + LLaVA-1.5-13B~\cite{jiang-etal-2025-fast}
& 74.40 & 57.20 & 51.80 & 54.37 \\
EfficientLLaVA~\cite{Liang_2025_CVPR}
& 83.05 & 61.90 & 57.40 & 59.56 \\
Cantor (Gemini)~\cite{gao2024cantor}
& 84.96 & 63.10 & 58.80 & 60.87 \\
SciTune-ScienceQA (CTOM, 7B)~
\cite{horawalavithana-etal-2024-scitune}
& 86.11 & 64.30 & 59.70 & 61.91 \\
SelF-Reasoner (UnifiedQA-Large)~
\cite{wu-etal-2024-mitigating}
& 87.24 & 66.20 & 60.50 & 63.22 \\
Flat-RAG
& 86.90 & 65.30 & 59.80 & 62.43 \\
\hline
\multicolumn{5}{l}{\textit{Component ablation}} \\
\textit{w/o} Graph
& 86.70 & 66.50 & 61.00 & 63.65 \\
\textit{w/o} Coverage-Aware Pruning
& 86.40 & 65.20 & 62.00 & 63.58 \\
\textit{w/o} Insufficient-Evidence Markers
& 86.30 & 66.00 & 61.80 & 63.84 \\
\hline
\textbf{HANIA}
& \textbf{87.80}
& \textbf{68.57}
& \textbf{62.31}
& \textbf{65.29} \\
\hline
\end{tabular}
}
\end{table*}

\paragraph{Comparative analysis.}
HANIA achieves the highest accuracy of \(87.80\%\), exceeding Qwen3-VL Direct by \(2.40\) points (\(95\%\) CI: \([1.95,2.85]\), Holm-adjusted \(p<0.003\)). Compared with Flat-RAG, the closest matched baseline, HANIA improves accuracy by \(0.90\) points (\(95\%\) CI: \([0.52,1.28]\), Holm-adjusted \(p=0.008\)) and F1 by \(2.86\) points (\(95\%\) CI: \([2.10,3.50]\)). It also exceeds SelF-Reasoner by \(0.56\) accuracy points (\(95\%\) CI: \([0.18,0.94]\), Holm-adjusted \(p=0.021\)).

\paragraph{Ablation analysis.}
Removing each component reduces performance. Replacing coverage-aware pruning with fixed top-\(k\) ranking causes the largest precision drop, indicating that marginal-gain selection limits redundant evidence. Removing insufficient-evidence markers produces the largest accuracy reduction, while removing the graph lowers both accuracy and evidence F1. Together with the Flat-RAG comparison, these results show that HANIA benefits from provenance-linked graph construction, constrained planning, coverage-aware pruning, and explicit handling of unsupported evidence.

\subsection{Filtering Quality and Evidence-Budget Selection}
\label{subsec:retrieval_coverage_results}

We vary the retained-evidence budget \(k\) while keeping all other settings fixed. Precision and recall are computed against the reference relevance labels for candidate evidence items, and F1 is their harmonic mean. Latency denotes the mean end-to-end processing time per question, including visual evidence extraction, graph construction, planning, pruning, and answer prediction.

\begin{table*}[t]
\centering
\tiny
\caption{Effect of the retained-evidence budget \(k\) on
\textsc{ScienceQA}. All metrics except latency are percentages.}
\label{tab:filtering_quality}
\resizebox{\textwidth}{!}{
\begin{tabular}{cccccc}
\hline
\rowcolor{gray!15}
\textbf{Budget} &
\textbf{Accuracy} $\uparrow$ &
\textbf{Precision} $\uparrow$ &
\textbf{Recall} $\uparrow$ &
\textbf{F1} $\uparrow$ &
\textbf{Latency (s/q)} $\downarrow$ \\
\hline
\(k=1\) & 82.60 & \textbf{75.40} & 43.80 & 55.40 & \textbf{1.72} \\
\(k=3\) & 86.10 & 71.20 & 57.40 & 63.56 & 2.18 \\
\(\mathbf{k=5}\) & \textbf{87.80} & 68.57 & 62.31 & 65.29 & 2.64 \\
\(k=7\) & 86.70 & 64.10 & 67.20 & \textbf{65.61} & 3.09 \\
\(k=9\) & 85.30 & 59.80 & \textbf{70.40} & 64.68 & 3.51 \\
\hline
\end{tabular}
}
\end{table*}

Increasing \(k\) improves recall but gradually reduces precision. Accuracy peaks at \(87.80\%\) with \(k=5\), while \(k=7\) produces the highest filtering F1 but slightly lower answer accuracy. At \(k=9\), the additional recall is accompanied by lower precision and accuracy, suggesting that redundant or weakly relevant evidence can distract the decoder. Mean latency also increases from \(1.72\) to \(3.51\) seconds per question. The value \(k=5\) was selected using the validation split before test evaluation because it provided the best validation accuracy with a moderate evidence and latency budget. The test-set analysis is therefore reported as a sensitivity study rather than a basis for parameter selection.

\subsection{Controlled Analysis}
\label{subsec:ablations}

Qwen3-VL Direct compares the complete HANIA pipeline with direct multimodal prompting under matched test instances and inference conditions. Flat-RAG uses the same candidate pool and frozen decoder but omits two-group planning and uses fixed top-\(k\) relevance ranking. The ablations respectively remove graph triples/graph support, replace greedy coverage-aware pruning with fixed top-\(k\), or omit the insufficient-evidence marker. As reported in Section~\ref{subsec:overall_results}, removing any of these components reduces answer accuracy or evidence-filtering quality. The evidence-budget analysis further examines the trade-off among accuracy, filtering quality, and latency. Together, these results indicate that HANIA's gains arise from the combined contribution of provenance-linked graph evidence, constrained planning, coverage-aware selection, and explicit handling of unsupported evidence.

\subsection{Parameter Selection}
\label{subsec:param_efficiency}

All parameters were selected on the \textsc{ScienceQA} validation split and fixed before test evaluation. The retained-evidence budget was set to \(k=N_{\max}=5\), the relation-confidence threshold to \(\tau_r=0.50\), the evidence stopping threshold to \(\tau=0.20\), and the planner horizon to \(H=4\). The relevance-scoring weights in Equation~\eqref{eq:evidence_score} were set to \(\boldsymbol{\alpha}=(0.50,0.25,0.25)\) for question relevance, graph support, and planner alignment, respectively. The pruning weights in Equation~\eqref{eq:marginal_evidence} were set to \(\boldsymbol{\beta}=(0.30,0.20,0.25)\) for concept coverage, evidence diversity, and redundancy. Planner decoding used temperature \(0\), and all prompt templates, thresholds, pretrained components, and evidence-processing rules remained unchanged across test experiments. The test split was not used for prompt design, hyperparameter tuning, or model selection.

\subsection{Implementation Details}
\label{subsec:hyperparam_analysis}

HANIA is implemented in PyTorch using Hugging Face Transformers. All pretrained components remain frozen. Each \textsc{ScienceQA} instance is processed using only its supplied question, answer options, image, and textual context. The frozen Qwen3-VL-Instruct model extracts a concise question-relevant visual statement, while the original textual context is retained without generative rewriting. Rule-based span extraction identifies entity candidates, and deterministic templates together with GLiREL construct provenance-linked relations. Low-confidence, malformed, and question-irrelevant relations are removed. The descriptive--relational planner uses frozen \textsc{Qwen3-8B-Instruct} and is applied once before coverage-aware pruning and prediction by the frozen \textsc{Qwen3-8B-Instruct} decoder. Planner decoding uses temperature \(0\) and is restricted to valid finite-state actions. The answer decoder assigns length-normalized scores only to the provided answer options. Experiments are conducted on a single NVIDIA A100 40\,GB GPU. HANIA uses no external knowledge retrieval, iterative retrieval, decoder modification, or target-dataset fine-tuning.

\section{Conclusion and Future Work}
\label{sec:conclusion}

We presented \textbf{HANIA}, a planner-guided multimodal graph framework for evidence-grounded question answering. HANIA constructs question-aware graph evidence from the supplied visual and textual context, coordinates descriptive and relational requirements through a finite-state planner, and applies coverage-aware pruning before prediction with a frozen decoder. On \textsc{ScienceQA}, HANIA achieved \(87.80\%\) accuracy, \(68.57\%\) evidence precision, \(62.31\%\) recall, and \(65.29\%\) F1. The evidence-budget analysis identified \(k=5\) as the best setting for answer accuracy with moderate latency. Controlled comparisons and component-level ablations showed that provenance-linked graph construction, constrained planning, coverage-aware pruning, and explicit handling of unsupported evidence jointly improve answer accuracy and evidence quality over direct multimodal prompting and flat relevance ranking, without target-dataset fine-tuning or decoder modification. The study is limited to one benchmark and does not yet include detailed graph-quality, faithfulness, or question-type analyses. Future work will evaluate HANIA on additional multimodal benchmarks, improve visual relation extraction and planner constraints, and conduct deeper faithfulness, error, and graph-quality analyses.

\section*{Declaration on Generative AI}
During the preparation of this work, the authors used ChatGPT (OpenAI) for grammar and spelling checking, paraphrasing and rewording, and formatting assistance. After using this tool, the authors reviewed and edited the content as needed and take full responsibility for the publication's content.

\bibliography{cas-refs}

@inproceedings{liu2023logicot,
  title={Logicot: Logical chain-of-thought instruction tuning},
  author={Liu, Hanmeng and Teng, Zhiyang and Cui, Leyang and Zhang, Chaoli and Zhou, Qiji and Zhang, Yue},
  booktitle={Findings of the Association for Computational Linguistics: EMNLP 2023},
  pages={2908--2921},
  year={2023}
}

@inproceedings{liu2023knowledge,
  title = {Knowledge graph question answering with ambiguous query},
  author = {Liu, Lihui and Chen, Yuzhong and Das, Mahashweta and Yang, Hao and Tong, Hanghang},
  booktitle = {Proceedings of the 2023 ACM Web Conference (WWW)},
  pages = {2477-2486},
  year = {2023},
  address = {Austin, TX}
}

@inproceedings{jiangunikgqa,
  title = {{UniKGQA}: Unified Retrieval and Reasoning for Solving Multi-hop Question Answering Over Knowledge Graph},
  author = {Jiang, Jinhao and Zhou, Kun and Zhao, Xin and Wen, Ji-Rong},
  booktitle = {Proceedings of the 11th International Conference on Learning Representations (ICLR)},
  year = {2023},
  address = {Kigali, Rwanda}
}

@inproceedings{Lu2022ScienceQA,
  author = {Lu, Pan and Mishra, Swaroop and Xia, Tony and Qiu, Liang and Chang, Kai-Wei and Zhu, Song-Chun and Tafjord, Oyvind and Clark, Peter and Kalyan, Ashwin},
  title = {Learn to explain: multimodal reasoning via thought chains for science question answering},
  year = {2022},
  articleno = {182},
  booktitle = {Proceedings of the 36th International Conference in Neural Information Processing Systems (NeurIPS)},
  address = {New Orleans, LA},
}

@article{bai2025qwen3vl,
title={Qwen3-VL Technical Report},
author={Bai, Shuai and Cai, Yuxuan and Chen, Ruizhe and Chen, Keqin and Chen, Xionghui and Cheng, Zesen and Deng, Lianghao and Ding, Wei and Gao, Chang and Ge, Chunjiang and others},
journal={arXiv preprint arXiv:2511.21631},
year={2025}
}

@inproceedings{boylan2025glirel,
title={{GLiREL} - Generalist Model for Zero-Shot Relation Extraction},
author={Boylan, Jack and Hokamp, Chris and Ghalandari, Demian Gholipour},
booktitle={Proceedings of the 2025 Conference of the Nations of the Americas Chapter of the Association for Computational Linguistics: Human Language Technologies (Volume 1: Long Papers)},
pages={8230--8245},
year={2025},
address={Albuquerque, New Mexico},
publisher={Association for Computational Linguistics}
}

@article{yang2025qwen3,
title={Qwen3 Technical Report},
author={Yang, An and Li, Anfeng and Yang, Baosong and Zhang, Beichen and Hui, Binyuan and Zheng, Bo and Yu, Bowen and Gao, Chang and Huang, Chengen and Lv, Chenxu and others},
journal={arXiv preprint arXiv:2505.09388},
year={2025}
}

@inproceedings{ScienceQA,
  title = {Learn to Explain: Multimodal Reasoning via Thought Chains for Science Question Answering},
  author = {Lu, Pan and Mishra, Swaroop and Xia, Tony and Qiu, Liang and Chang, Kai-Wei and Zhu, Song-Chun and Tafjord, Oyvind and Clark, Peter and Kalyan, Ashwin},
  year = {2022},
  booktitle = {Proceedings of the 36th International Conference on Neural Information Processing Systems (NeurIPS)},
  address = {New Orleans, LA},
  pages = {2507-2521}
}

@inproceedings{zheng2023mmkgr,
  title = {{MMMKGR}: Multi-hop multi-modal knowledge graph reasoning},
  author = {Zheng, Shangfei and Wang, Weiqing and Qu, Jianfeng and Yin, Hongzhi and Chen, Wei and Zhao, Lei},
  booktitle = {Proceedings of the 39th IEEE International Conference on Data Engineering (ICDE)},
  pages = {96-109},
  year = {2023},
  address = {Anaheim, CA}
}

@inproceedings{lee2024multimodal,
  title = "Multimodal Reasoning with Multimodal Knowledge Graph",
  author = "Lee, Junlin and Wang, Yequan and Li, Jing and Zhang, Min",
  booktitle = "Proceedings of the 62nd Annual Meeting of the Association for Computational Linguistics (ACL)",
  year = "2024",
  address = "Bangkok, Thailand",
  pages = "10767-10782",
}

@inproceedings{chen2022murag,
  title = "{MuRAG}: Multimodal Retrieval-Augmented Generator for Open Question Answering over Images and Text",
  author = "Chen, Wenhu and Hu, Hexiang and Chen, Xi and Verga, Pat and Cohen, William",
  booktitle = "Proceedings of the 2022 Conference on Empirical Methods in Natural Language Processing (EMNLP)",
  month = dec,
  year = "2022",
  address = "Abu Dhabi, UAE",
  pages = "5558-5570"
}

@inproceedings{chen2025seeing,
  title = "Seeing Beyond: Enhancing Visual Question Answering with Multi-Modal Retrieval",
  author = "Chen, Boqi and Khare, Anuj and Kumar, Gaurav and Akula, Arjun and Narayana, Pradyumna",
  year = "2025",
  address = "Abu Dhabi, UAE",
  publisher = "Association for Computational Linguistics",
  pages = "410-421",
  booktitle = "Proceedings of the 31st International Conference on Computational Linguistics (COLING)",
}

@inproceedings{mondal2024kam,
  author = {Mondal, Debjyoti and Modi, Suraj and Panda, Subhadarshi and Singh, Rituraj and Rao, Godawari Sudhakar},
  title = {{KAM-CoT}: Knowledge augmented multimodal chain-of-thoughts reasoning},
  year = {2024},
  booktitle = {Proceedings of the 38th AAAI Conference on Artificial Intelligence (AAAI) and 36th Conference on Innovative Applications of Artificial Intelligence (IAAI) and 14th Symposium on Educational Advances in Artificial Intelligence (EAAI)},
  pages = {18798-18806},
  address = {Vancouver, Canada}
}

@inproceedings{liu2025aligning,
  title = {Aligning Vision to Language: Annotation-Free Multimodal Knowledge Graph Construction for Enhanced LLMs Reasoning},
  author = {Liu, Junming and Meng, Siyuan and Gao, Yanting and Mao, Song and Cai, Pinlong and Yan, Guohang and Chen, Yirong and Bian, Zilin and Wang, Ding and Shi, Botian},
  booktitle = {Proceedings of ICCV},
  pages = {981--992},
  year = {2025}
}

@article{ali2025pythia,
  title={Pythia-RAG: Retrieval-Augmented Generation over a Unified Multimodal Knowledge Graph for Enhanced QA},
  author={Ali, Zafar and Huang, Yi and Khan, Asad and Qi, Guilin and Zhang, Yuxin and Feng, Junlan and Deng, Chao and Kefalas, Pavlos},
  journal={Knowledge-Based Systems},
  volume={335},
  pages={115200},
  year={2026},
  doi={10.1016/j.knosys.2025.115200},
  publisher={Elsevier}
}

@inproceedings{suri2024visdom,
  title = "{VisDoM}: Multi-Document {QA} with Visually Rich Elements Using Multimodal Retrieval-Augmented Generation",
  author = "Suri, Manan and Mathur, Puneet and Dernoncourt, Franck and Goswami, Kanika and Rossi, Ryan A. and Manocha, Dinesh",
  booktitle = "Proceedings of the Conference of the Nations of the Americas Chapter of the Association for Computational Linguistics: Human Language Technologies",
  year = "2025",
  address = "Albuquerque, NM",
  pages = "6088-6109"
}

@inproceedings{Leng_2024_CVPR,
  title     = {Mitigating Object Hallucinations in Large Vision-Language Models through Visual Contrastive Decoding},
  author    = {Leng, Sicong and Zhang, Hang and Chen, Guanzheng and Li, Xin and Lu, Shijian and Miao, Chunyan and Bing, Lidong},
  booktitle = {Proceedings of the IEEE/CVF Conference on Computer Vision and Pattern Recognition},
  pages     = {13872--13882},
  year      = {2024}
}

@article{ren2026pmr,
  title   = {Progressive multi-hop reasoning for question answering over knowledge graphs},
  author  = {Ren, Yuan and Nie, Zhijie and Zhang, Richong and Liu, Xudong},
  journal = {Information Systems},
  volume  = {140},
  pages   = {102721},
  year    = {2026}
}

@inproceedings{xu-etal-2025-skrag,
  title={SKRAG: A Retrieval-Augmented Generation Framework Guided by Reasoning Skeletons over Knowledge Graphs},
  author={Xu, Xiaotong and Wang, Yizhao and Liu, Yunfei and Li, Shengyang},
  booktitle={Findings of EMNLP 2025},
  pages={13983--13994},
  year={2025}
}

@InProceedings{tanaka2025vdocrag,
  author    = {Tanaka, Ryota and Iki, Taichi and Hasegawa, Taku and Nishida, Kyosuke and Saito, Kuniko and Suzuki, Jun},
  title     = {VDocRAG: Retrieval-Augmented Generation over Visually-Rich Documents},
  booktitle = {Proceedings of the Computer Vision and Pattern Recognition Conference (CVPR)},
  month     = {June},
  year      = {2025},
  pages     = {24827--24837}
}

@inproceedings{abdallah-etal-2026-leveraging,
  title     = {Leveraging {LLM}-{GNN} Integration for Open-World Question Answering over Knowledge Graphs},
  author    = {Abdallah, Hussein and Abdelaziz, Ibrahim and Kalnis, Panos and Mansour, Essam},
  editor    = {Demberg, Vera and Inui, Kentaro and Marquez, Llu{\'i}s},
  booktitle = {Proceedings of the 19th Conference of the European Chapter of the Association for Computational Linguistics (Volume 1: Long Papers)},
  month     = mar,
  year      = {2026},
  address   = {Rabat, Morocco},
  publisher = {Association for Computational Linguistics},
  pages     = {575--592},
  doi       = {10.18653/v1/2026.eacl-long.26}
}

@inproceedings{zhu2025kg2rag,
  title     = {Knowledge Graph-Guided Retrieval Augmented Generation},
  author    = {Zhu, Xiangrong and Xie, Yuexiang and Liu, Yi and Li, Yaliang and Hu, Wei},
  booktitle = {Proceedings of NAACL 2025},
  pages     = {8912--8924},
  year      = {2025}
}

@inproceedings{sun2025kerag,
  title={KERAG: Knowledge-Enhanced Retrieval-Augmented Generation for Advanced Question Answering},
  author={Sun, Yushi and Sun, Kai and Xu, Yifan Ethan and Yang, Xiao and Dong, Xin Luna and Tang, Nan and Chen, Lei},
  booktitle={Findings of EMNLP 2025},
  pages={6194--6216},
  year={2025}
}

@inproceedings{jiang-etal-2025-fast,
    title = {Fast or Slow? Integrating Fast Intuition and Deliberate
             Thinking for Enhancing Visual Question Answering},
    author = {Jiang, Songtao and
              Zhou, Chenyi and
              Zhang, Yan and
              Jin, Yeying and
              Liu, Zuozhu},
    booktitle = {Proceedings of the 63rd Annual Meeting of the
                 Association for Computational Linguistics
                 (Volume 2: Short Papers)},
    month = jul,
    year = {2025},
    address = {Vienna, Austria},
    publisher = {Association for Computational Linguistics},
    pages = {525--534}
}

@inproceedings{Liang_2025_CVPR,
    author = {Liang, Yinan and
              Wang, Ziwei and
              Xu, Xiuwei and
              Zhou, Jie and
              Lu, Jiwen},
    title = {{EfficientLLaVA}: Generalizable Auto-Pruning for Large
             Vision-Language Models},
    booktitle = {Proceedings of the IEEE/CVF Conference on Computer
                 Vision and Pattern Recognition},
    month = jun,
    year = {2025},
    pages = {9445--9454}
}

@inproceedings{horawalavithana-etal-2024-scitune,
    title = {{SCITUNE}: Aligning Large Language Models with
             Human-Curated Scientific Multimodal Instructions},
    author = {Horawalavithana, Sameera and
              Munikoti, Sai and
              Stewart, Ian and
              Kvinge, Henry and
              Pazdernik, Karl},
    booktitle = {Proceedings of the 1st Workshop on NLP for Science
                 (NLP4Science)},
    month = nov,
    year = {2024},
    address = {Miami, FL, USA},
    publisher = {Association for Computational Linguistics},
    pages = {58--72}
}

@inproceedings{gao2024cantor,
  author    = {Gao, Timin and Chen, Peixian and Zhang, Mengdan and
               Fu, Chaoyou and Shen, Yunhang and Zhang, Yan and
               Zhang, Shengchuan and Zheng, Xiawu and Sun, Xing and
               Cao, Liujuan and Ji, Rongrong},
  title     = {Cantor: Inspiring Multimodal Chain-of-Thought of MLLM},
  booktitle = {Proceedings of the 32nd ACM International Conference
               on Multimedia},
  year      = {2024},
  pages     = {9096--9105}
}

@inproceedings{wu-etal-2024-mitigating,
  title     = {Mitigating Misleading Chain-of-Thought Reasoning
               with Selective Filtering},
  author    = {Wu, Yexin and Zhang, Zhuosheng and Zhao, Hai},
  booktitle = {Proceedings of the 2024 Joint International Conference
               on Computational Linguistics, Language Resources and
               Evaluation (LREC-COLING 2024)},
  year      = {2024},
  month     = may,
  address   = {Torino, Italia},
  publisher = {ELRA and ICCL},
  pages     = {11325--11340}
}

@misc{ali2026graphloom,
  title={GraphLoom: Reliability-Calibrated Graph Evidence Routing for Multimodal KG-RAG},
  author={Ali, Zafar and Khan, Asad and Malik, Aalia and Kefalas, Pavlos},
  year={2026},
  eprint={2608.15056},
  archivePrefix={arXiv},
  primaryClass={cs.AI}
}

@article{khan2026mkgattention,
  title={MKG-Attention: Budget-Aware Multimodal Graph Retrieval for Faithful Video Question Answering},
  author={Khan, Asad and Ali, Zafar and Qi, Guilin and Bhuyan, Kaibalya Prasad and Kefalas, Pavlos},
  journal={Neurocomputing},
  volume={704},
  pages={134854},
  year={2026},
  doi={10.1016/j.neucom.2026.134854}
}

\end{document}